\pdfoutput=1

\documentclass[11pt]{article}

\usepackage[]{acl}

\usepackage{times}
\usepackage{latexsym}
\usepackage{arydshln}
\usepackage[T1]{fontenc}

\usepackage[utf8]{inputenc}

\usepackage{microtype}

\usepackage{tikz}
\usepackage{pgfplots} 
\pgfplotsset{compat=1.12}
\usepackage{pgfplotstable}
\usepackage{graphicx}
\usepackage{tabularx}
\usepackage{booktabs}
\usepackage{multirow}
\usepackage{array}
\usepackage{makecell}
\usepackage{xcolor}
\usepackage{xstring}
\usepackage{color}
\usepackage{subcaption}
\usepackage{colortbl}

\usepackage{hyperref}
 \definecolor{darkblue}{rgb}{0, 0, 0.5}
  \hypersetup{colorlinks=true, citecolor=darkblue, linkcolor=darkblue, urlcolor=darkblue}

\newcommand\blfootnote[1]{%
  \begingroup
  \renewcommand\thefootnote{}\footnote{#1}%
  \addtocounter{footnote}{-1}%
  \endgroup
}

\title{Distilling Text Style Transfer With Self-Explanation From LLMs}

\author{Chiyu Zhang$^{1,2,\star}$ ~~~Honglong Cai$^2$ ~~~Yuezhang (Music) Li$^2$ ~~~Yuexin Wu$^2$ \\ \textbf{\large Le Hou$^2$} ~~~\textbf{ \large Muhammad Abdul-Mageed$^{1,3}$}  \\
$^1$~The University of British Columbia ~~~$^2$~Google \\ $^3$Department of NLP \& ML, MBZUAI \\
\texttt{chiyuzh@mail.ubc.ca}, \texttt{\{honglongcai, lyzmuisc\}@google.com}}

\begin{document}

\maketitle

\begin{abstract}
Text Style Transfer (TST) seeks to alter the style of text while retaining its core content. Given the constraints of limited parallel datasets for TST, we propose CoTeX, a framework that leverages large language models (LLMs) alongside chain-of-thought (CoT) prompting to facilitate TST. CoTeX distills the complex rewriting and reasoning capabilities of LLMs into more streamlined models capable of working with both non-parallel and parallel data. Through experimentation across four TST datasets, CoTeX is shown to surpass traditional supervised fine-tuning and knowledge distillation methods, particularly in low-resource settings. We conduct a comprehensive evaluation, comparing CoTeX against current unsupervised, supervised, in-context learning (ICL) techniques, and instruction-tuned LLMs. Furthermore, CoTeX distinguishes itself by offering transparent explanations for its style transfer process.
\end{abstract}

 ~\blfootnote{ $^{\star}${Work done during internship at Google.}}
 
\begin{figure*}[t]
\centering
\includegraphics[width=\linewidth]{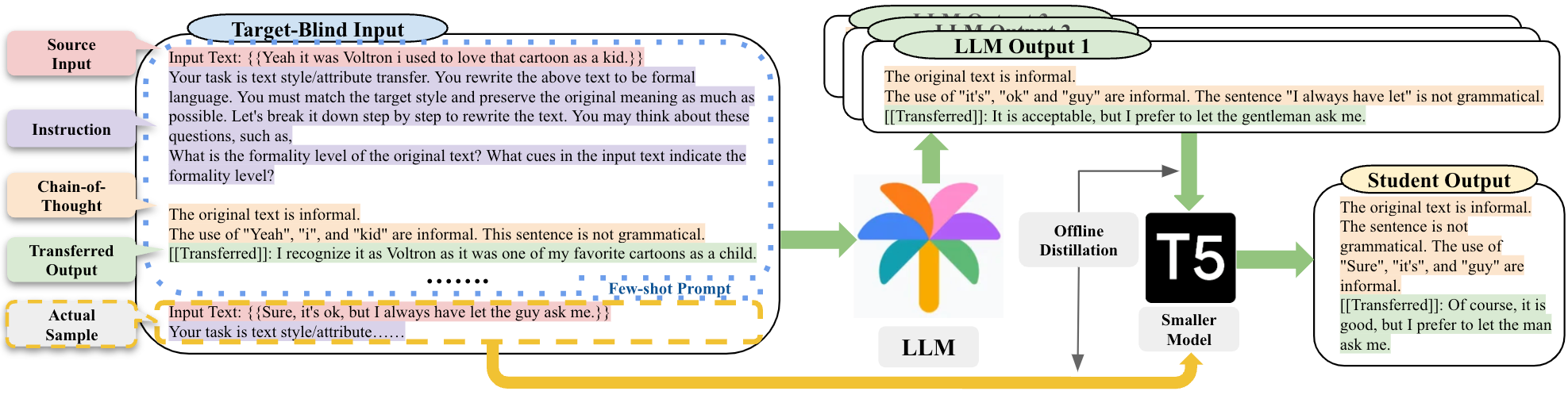}

  \caption{Overview of CoTeX framework. We use few-shot CoT prompting to generate reasoning paths and transferred texts from an LLM and then train a smaller task-specific model with generated data.}
  \label{fig:generation_example}
\end{figure*}

\section{Introduction}
TST aims to rephrase a source text $s$ with the desired style $\tau$ while preserving its core meaning and ensuring fluency of the generated text $t$~\cite{di-2022-deep}.
The term ``style" can encompass the personal characteristics of an author, such as age, and pragmatic use like formality or toxicity. To develop TST systems using supervised methods, several human-annotated datasets have emerged~\cite{rao-tetreault-2018-dear}. For instance, \citet{rao-tetreault-2018-dear} introduced a corpus for formality style transfer, transforming informal language to its formal counterpart and vice versa. 
Nonetheless, supervised parallel data, crucial for training deep neural networks, is scarce and costly to obtain. Hence, unsupervised methodologies~\cite{shen-2017-style, liu2021non} have been proposed to manage stylistic attributes without relying on parallel data. \citet{liu-2022-semisupervised} and \citet{zhang-2020-parallel} create pseudo-parallel data from unlabeled samples via diverse data augmentation with task-specific knowledge.
Works by \citet{gong-2019-reinforcement, wang-2019-controllable, reid-2021-lewis} employ an auxiliary style classifier to steer the transfer direction. Meanwhile, \citet{krishna-2020-reformulating} and \citet{hallinan-2023-detoxifying} deploy multiple style-specific models to produce various styles individually. Of late, LLMs have demonstrated exceptional prowess across diverse NLP tasks. Studies like \citet{reif-2022-recipe, pu-2023-chatgpt} have found that extremely large LMs, with over 100B parameters, are adept at TST with ICL. Drawing from these findings, our paper uses LLMs to generate pseudo-parallel data and distills the TST skills of the LLM into a compact student model. Moreover, we enhance distillation and efficiency using CoT prompting.

LLMs have demonstrated impressive performance across various tasks and reasoning capabilities. CoT prompting~\cite{wei-2022-chain} is a promising technique that extracts these reasoning skills and enhances accuracy in target tasks. However, deploying these enormous LLMs poses computational and practical challenges. Recent studies~\cite{huang-2022-large, wang--2023-scott, hsieh-2023-distilling} have thus turned to offline knowledge distillation (KD)~\cite{hinton-2015-distilling} to condense these reasoning capabilities into a smaller model. Using CoT rationales can also increase distillation efficiency with less data~\cite{li-2022-explanations, shridhar-2023-distilling}. Concurrently, \citet{saakyan-2023-iclef} examine CoT prompting combined with domain expert feedback for improved formality transfer. Nevertheless, the potential of CoT prompting and KD to enrich a broader range of TST tasks remains underexplored. 

In this paper, we present CoTeX framework, using CoT prompting to improve TST. It identifies cues for TST and clarifies the rewriting process ($\S$~\ref{sec:method}). We then distill the reasoning and style transfer abilities of LLMs into compact models. We exploit CoT prompting to enhance TST, applicable to scenarios both with and without parallel data, and show the effectiveness of CoTeX in low-resource settings. 
Our \textbf{\textit{primary findings}} include: 
\textbf{(1)} Our target-blind CoTeX (CoTeX-TB) substantially boosts data efficiency for training smaller student models for TST. 
\textbf{(2)} The target-aware CoTeX (CoTeX-TA) consistently outperforms SFT and conventional KD across various datasets and training data sizes. 
\textbf{(3)} Our CoTeX-TB outperforms state-of-the-art (SoTA) unsupervised and ICL methods on three TST datasets.
\textbf{(4)} Leveraging CoT rationales, our distilled student models can elucidate the rewriting procedure.

\section{Method}\label{sec:method}

\subsection{Data Generation}
We employ CoT combined with instruction prompting to extract rationales from LLMs regarding the TST process. We have two different settings (target-blind and target-aware) to generate rationales. 

\paragraph{Target-Blind (TB).} We first explore our method in the target-blind setting where we only give a source text and the name of the desired target style. This setting can be adaptable to a broader range of style transfer directions. As shown in the left side of Figure~\ref{fig:generation_example}, each input example is constructed using an instruction template, $p_{tb}$. This template encompasses a source input $s_i$, a task instruction, the target style $\tau$, and a CoT trigger phrase: ``Let's break down the rewriting process step by step.'' LLM is tasked with producing the CoT, $c_i$, pertaining to the text rewriting process and the resultant transferred text, $\hat{t_i}$. To distinguish between the CoT and the transferred text, we instruct the model to initiate the transferred text on a new line, prefixed with a special token `[Transferred]:'. To facilitate the LLM's adherence to the desired output structure, we present $m$ examples created by humans as context before the actual input. In our implementation, we employ three manually crafted examples as few-shot prompts.

\begin{figure}[h]
\centering
\includegraphics[width=\linewidth]{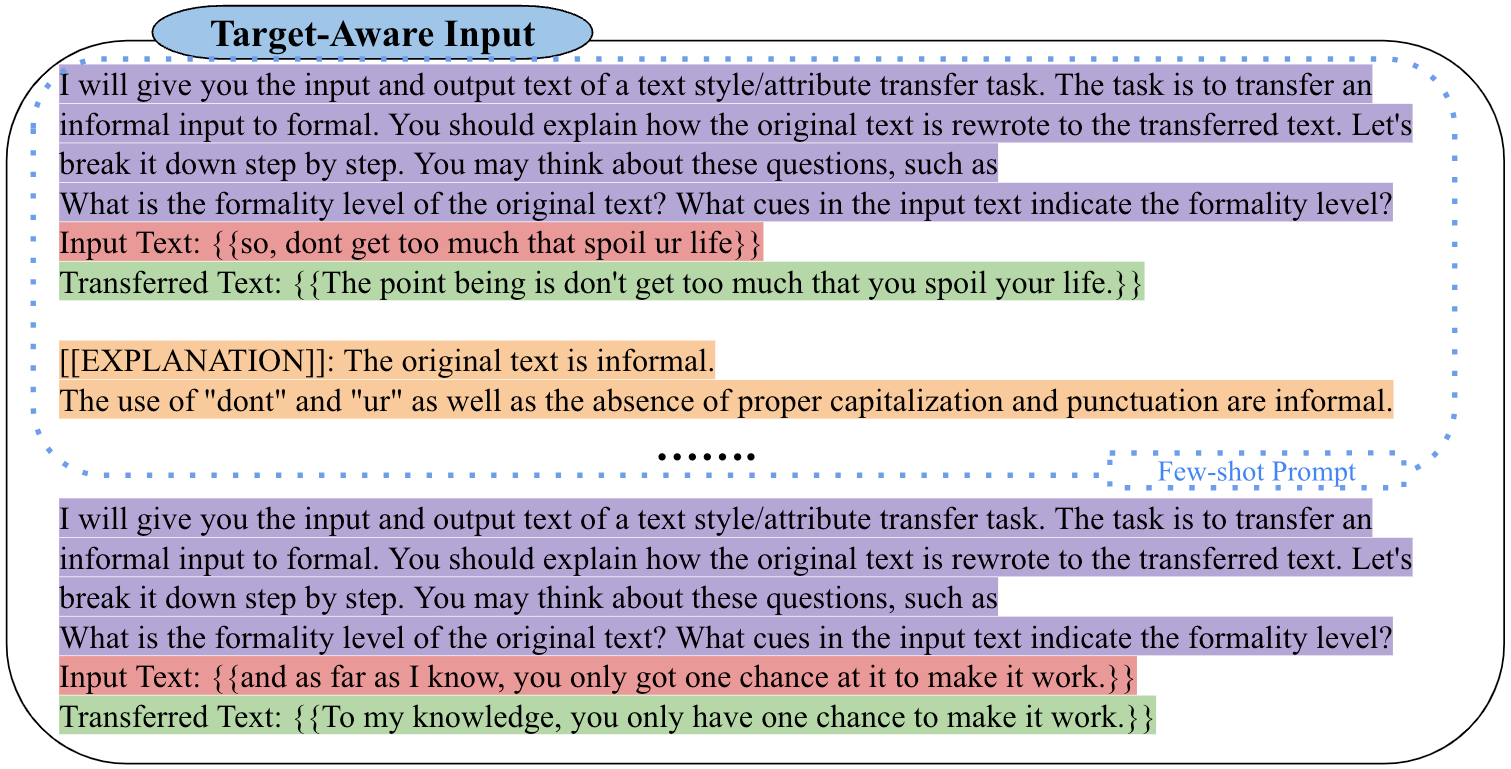}
  \caption{Few-shot chain-of-thought prompting for data generation with supervised data (target-aware setting). We use the few-shot prompts that include a few examples to guide LLM to generate desired outputs in a standard format.}
  \label{fig:supervised_prompt}
\end{figure}

\paragraph{Target-Aware (TA).} For datasets with supervised parallel data, we use the instruction template $p_{ta}$.\footnote{We manually tune both instruction templates and find the optimal templates used in this paper.} As Figure~\ref{fig:supervised_prompt} shows, this template $p_{ta}$ integrates a source text $s_i$, its corresponding human-annotated target text $t_i$, and the target style $\tau$. The LLM is then prompted to explain how $s_i$ is transformed into $t_i$, leading it to produce a CoT, $c_i$. This generated CoT is prefixed with a distinct token `[EXPLANATION]:'. To ensure LLMs produce outputs in the desired format, we also employ $m$ guiding examples, $m=3$ in our experiments.


\subsection{Training Student Models}
We leverage the LLM-generated data to finetune smaller, task-specific student models. For the data generated in the target-blind setting, we utilize the instruction template $p_{tb}$, which includes source text $s_i$ and target style $\tau$ as input. The corresponding supervision $\hat{y_i}$ for training is the generated CoT $c_i$ combined with the synthetic transferred text $\hat{t}_i$, i.e., $\hat{y_i}=c_i \oplus \hat{t}_i$. When employing the data generated from the target-aware setting, we also adopt the template $p_{tb}$. From this, we derive the generated CoT $c_i$ and merge it with the gold target text $t_i$. This composite then serves as the supervision $\hat{y_i}$ (i.e., $\hat{y_i}=c_i \oplus t_i$) for training a student model. A student model is trained with $\hat{y_i}$ employing the conventional cross-entropy loss.
\section{Experiments}\label{sec:experiment}

\subsection{Datasets and Metric}\label{subsec:dataset}
We employ four public datasets across three style transfer directions, chosen for their inclusion of human-annotated parallel data in both training and evaluation sets. This facilitates direct comparisons between different settings.

\noindent\textbf{Formality Transfer.} We use GYAFC dataset from \citet{rao-tetreault-2018-dear} and focus on the \textit{informal to formal language} transfer direction. GYAFC dataset includes two domains, Family \& Relationships (F\&R) and Entertainment \& Music (E\&M). 
~\textbf{Detoxification.} ParaDetox~\cite{logacheva-etal-2022-paradetox} is a parallel dataset for text detoxification. 
~\textbf{Shakespeare to Modern English.} \citet{xu2012paraphrasing} introduce a human-annotated dataset for translating text between William Shakespeare’s plays and their modernized versions. 



\noindent\textbf{Low-Resource Training.}
Our method offers advantages in low-resource settings, as the CoT is poised to enhance the learning efficiency of student models and bolster their generalizability. Thus, we create smaller training sets by randomly sampling training data, ranging from 1K to 20K.

\noindent\textbf{Evaluation Metric.}
We report BLEU, leveraging the Sacre-BLEU Python library~\cite{post-2018-call}, as main metric for evaluation.

\subsection{Model Comparison.} In low-resource settings, CoTeX is compared to \textbf{(1) SFT:} conventional supervised fine-tuning using parallel data, \textbf{(2) teacher LLM:} the teacher model evaluated on the Test set via few-shot ICL, i.e., using the three-shot prompt and template $p_{tb}$ described in Section~\ref{sec:method}, and \textbf{(3) Distill:} traditional offline knowledge distillation, which relies solely on LLM-generated pseudo-parallel data without a CoT path.

For comprehensive evaluations, CoTeX is further compared with \textbf{(1) Prompt\&Rank:} a SoTA in-context learning method for TST~\cite{suzgun-etal-2022-prompt}, and \textbf{(2) instruction-tuned LLMs:} open-source LLMs assessed through three-shot ICL using the same prompt and template described in Section~\ref{sec:method}; these LLMs include Alpaca 7B~\cite{taori-2023-alpaca}, Vicuna 7B~\cite{chiang-2023-vicuna}, LLaMA2-Chat 7B~\cite{touvron-2023-llama2}, and FlanT5-XL~\cite{chung-2022-flant5} (with 3B parameters). Additionally, for each dataset, comparisons are made with existing \textit{dataset-specific} unsupervised and supervised methods. \textbf{Unsupervised} methods include DualRL~\cite{luo-2019-dual}, STRAP~\cite{krishna-2020-reformulating}, DLS~\cite{he-2020-probabilistic}, and TSST~\cite{xiao-2021-transductive} for formality transfer; Mask\&Infill~\cite{wu-2019-mask} and CondBERT~\cite{dale-etal-2021-text} for detoxification; and STRAP and TSST for modernizing Shakespearean text. \textbf{Supervised} methods include Multi-NMT~\cite{niu-2018-multitask}, GPT-CAT~\cite{wang-2019-harnessing}, and SemiFST~\cite{liu-2022-semisupervised} for formality transfer; ParaDetox~\cite{logacheva-etal-2022-study} for detoxification; and PointerS2S~\cite{jhamtani-etal-2017-shakespearizing} for modernizing Shakespearean text.



\subsection{Implementation}\label{subsec:implementation}
We employ PaLM2 Unicorn~\cite{anil-2023-palm2} as our LLM for data generation. In the target-blind setting, we generate a CoT path and a transferred text.\footnote{Our ancillary study also examines the generation of multiple pairs of CoT paths and transferred text.} For the target-aware approach, we solely produce a CoT path. Both approaches use a temperature of 0.7. Afterward, we finetune a T5-large model (with 770M parameters)~\cite{raffel-2020-exploring} with the curated dataset.\footnote{We provide a concise experiment of using T5-XL model in Appendix~\ref{sec:t5xl}.} We finetune T5 for 2,000 steps with a learning rate of $1e-3$ and batch size of 128. We evaluate validation performance every 16 steps and report test result of the best step.\footnote{More details about hyperparameters are in Appendix~\ref{sec:app_hyperparameter}.}

\subsection{Hyperparameter for Training Student Model}\label{sec:app_hyperparameter}
We set the maximal input and output sequence lengths to 512 and 256, respectively. 
To optimize the T5 model's finetuning, we search both the learning rate and batch size within specified search spaces: $lr \in \{1e-3, 5e-4, 1e-5\}$ and batch size $\in \{32, 64, 128\}$. We undertake hyperparameter tuning using formality (F\&R) dataset. Based on the validation BLEU score, we identify the optimal hyperparameters are $lr=1e-3$ and batch size $=128$. We finetune T5 for 2,000 steps, evaluate performance on the validation set every 16 steps, and report the test performance on the best step. All T5 models are trained on four V3 TPUs.


\begin{figure}[t]
    \centering
    \includegraphics[width=\linewidth]{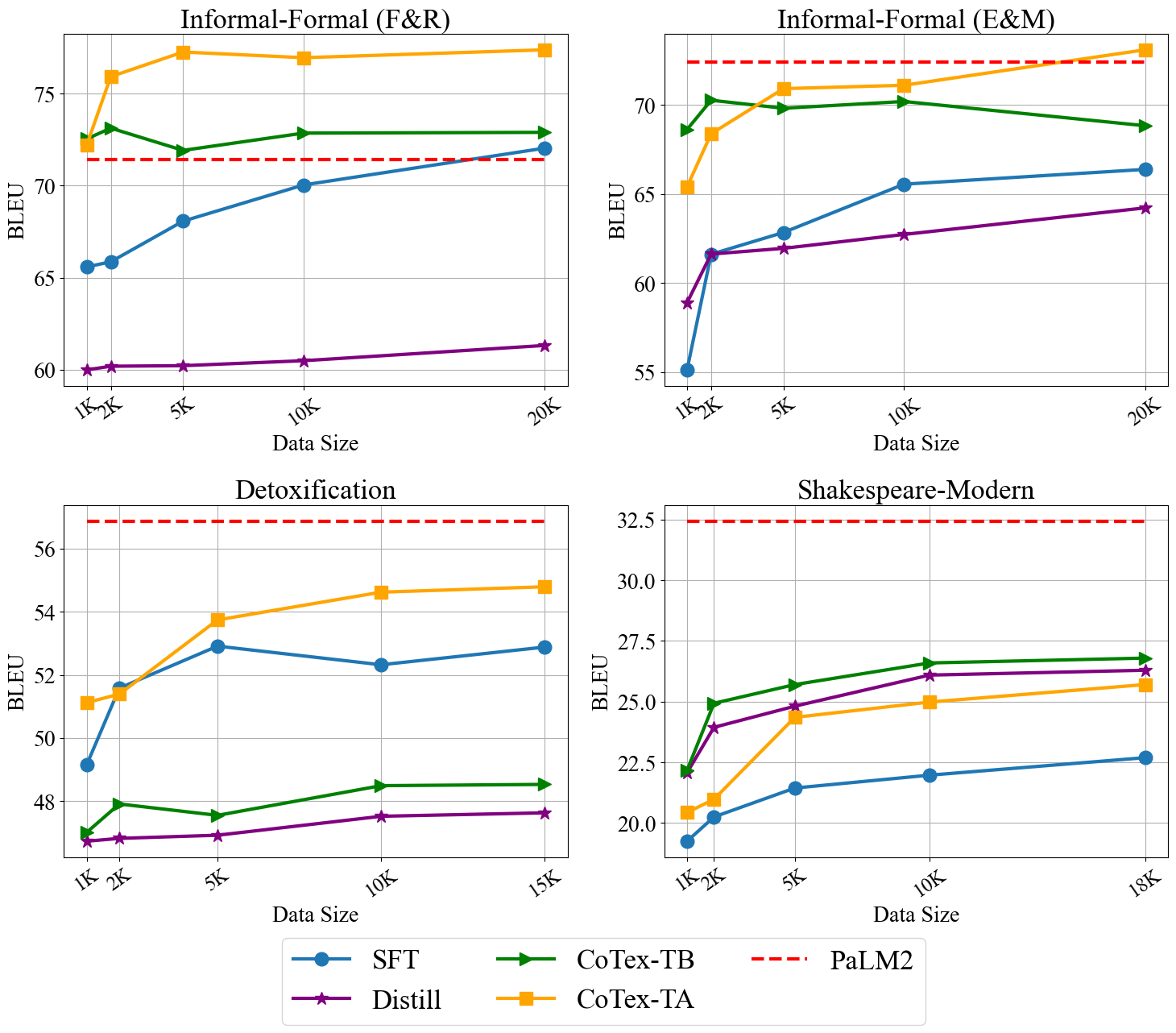}     
    \caption{Test results of low-resource settings.}
    \label{fig:results}
\end{figure}

\section{Results}\label{subsec:results}
We now present your experimental results. CoTeX-TB and CoTeX-TA denote models trained using datasets created through target-blind and target-aware methods, respectively.

\paragraph{Low-Resource Settings.} We first examine CoTeX's impact in low-resource context. Figure~\ref{fig:results} shows CoTeX's performance in both target-blind and target-aware settings across varying training data sizes. In both formality transfer datasets, CoTeX-TB outperforms SFT-T5 and Distill-T5. This advantage is noticeable with limited data, specifically under 10K. For instance, using just 1K samples from the informal-formal (E\&M) dataset, the BLEU scores for SFT, CoTeX-TB, and CoTeX-TA are 55.13, 68.62, and 65.40, respectively. 
We find that both CoTeX-TB and CoTeX-TA outperform or match the LLM's performance on the two formality datasets. In translating Shakespearean to modern English, CoTeX-TB exhibits significant superiority over SFT-T5 and Distill-T5 across all data sizes. We believe that such an enhancement can be attributed to the high quality of LLM generations. LLM with few-shot in-context learning obtains a BLEU score of 32.43. Though CoTeX-TB underperforms SFT on detoxification, CoTeX-TA still outperforms SFT in most data sizes. 

\begin{table}[]
\scriptsize
\centering
\begin{tabular}{@{}lll|ll@{}}
\toprule
                        & \multicolumn{1}{c}{\textbf{Method}} & \multicolumn{1}{c}{\textbf{BLEU}} & \multicolumn{1}{c}{\textbf{Method}} & \multicolumn{1}{c}{\textbf{BLEU}} \\ \midrule
                        & \multicolumn{2}{c|}{\textbf{Formality (F\&R)}}                                & \multicolumn{2}{c}{\textbf{Formality (E\&M)}}                                \\\midrule
\multirow{2}{*}{Unsup.} & DualRL               & 53.01                             & DLS       & 23.09                             \\
                        & TSST      & 60.99                             & STRAP   & 31.39                             \\\cdashline{1-5}
\multirow{5}{*}{ICL}    & Prompt\&Rank                         & 30.60                             & Prompt\&Rank                         & 30.96                             \\
                        & Alpaca                              & 41.85                             & Alpaca                              & 52.40                             \\
                        & Vicuna                              & 37.09                             & Vicuna                              & 46.47                             \\
                        & LLaMA2-C.                     & 19.62                             & LLaMA2-C.                      & 25.14                             \\
                        & FlanT5-XL                            & 55.70                             & FlanT5-XL                            & 42.58                             \\\cdashline{1-5}
\multirow{4}{*}{Sup.}   & Multi-NMT$^\dagger$          & 75.35                             & Multi-NMT$^\dagger$          & 72.01                             \\
                        & GPT-CAT$^\dagger$        & 77.26                             & GPT-CAT$^\dagger$        & 71.39                             \\
                        & SemiFST     & \textbf{80.32}                    & SemiFST     & \textbf{76.87}                    \\
                        & SFT (ours)                          & 77.12                             & SFT (ours)                          & 73.01                             \\\cdashline{1-5}
                    & Distill (ours)                      & 64.79                             & Distill (ours)                      & 64.31                             \\\cdashline{1-5}
                        & CoTex-TB                  & \underline{72.05}                         & CoTex-TB                  & \underline{71.70}                             \\
                        & CoTex-TA                   & 77.13                             & CoTex-TA                   & 74.65                             \\ \midrule
                        & \multicolumn{2}{c|}{\textbf{Detoxification}}                             & \multicolumn{2}{c}{\textbf{Modernizing Shake.}}              \\\midrule
\multirow{2}{*}{Unsup.} & Mask\&Infill$^*$            & 44.77                             & DLS        & 12.85                             \\
                        & CondBERT$^*$     & 48.89                             & STRAP   & 19.96                             \\\cdashline{1-5}
\multirow{5}{*}{ICL}    & Prompt\&Rank                         & 11.06                             & Prompt\&Rank                         & 20.87                             \\
                        & Alpaca                              & 24.32                             & Alpaca                              & 24.33                             \\
                        & Vicuna                              & 34.54                             & Vicuna                              & 17.76                             \\
                        & LLaMA2-C.                      & 14.65                             & LLaMA2-C.                      & 25.19                             \\
                        & FlanT5-XL                            & \underline{50.13}                             & FlanT5-XL                            & 21.55                             \\\cdashline{1-5}
\multirow{2}{*}{Sup.}                   & ParaDetox                           & 53.98                             & PointerS2S                                & \textbf{30.78}                             \\ 
                        & SFT (ours)                          & 52.88                             & SFT (ours)                          & 22.69                             \\\cdashline{1-5}
                    & Distill (ours)                      & 43.97                             & Distill (ours)                      & 22.88                             \\\cdashline{1-5}
                        & CoTex-TB                  & 48.53                             & CoTex-TB                  &  \underline{26.79}                    \\
                        & CoTex-TA                   & \textbf{54.79}                    & CoTex-TA                  & 25.70                             \\ \bottomrule
\end{tabular}
\caption{Comparing to previous methods. The best-performed method is in \textbf{bold}. The best method without utilizing a full parallel Train set is \underline{underscored}. \textbf{Unsup.:} unsupervised, \textbf{Sup.:} supervised, \textbf{$^\dagger$}: Take from~\citet{liu-2022-semisupervised}. \textbf{$^*$}: Utilize outputs from implementation of~\citet{logacheva-etal-2022-paradetox}.}\label{tab:all_res}
\end{table}

\paragraph{Utilizing the Full Dataset.} Training student models with CoTeX on all training samples of each dataset, we present comparative results in Tables~\ref{tab:all_res}.\footnote{CoTeX-TB setting utilizes the source text from training sample while keeping the target undisclosed.} 
Given that many unsupervised TST studies have not reported BLEU scores, we compute BLEU scores for their public outputs using our evaluation scripts to ensure a fair comparison. CoTeX-TB surpasses previous unsupervised methods, the SoTA ICL method Prompt\&Rank, and instruction-tuned LLMs across both domains within the formality transfer dataset.
Although CoTeX-TA does not exceed the performance of SoTA supervised methods, SemiFST, for formality transfer, it is noteworthy that our method does not depend on task-specific data augmentation strategies or knowledge, offering greater flexibility. 
In the detoxification task, our results are compared with the top-performing model from~\citet{logacheva-etal-2022-paradetox}. CoTeX-TA outperforms previous supervised methods, while CoTeX-TB falls slightly short of CondBERT, which employs additional style-conditional LMs for transfer control. FlanT5-XL, an instruction-tuned LLM, leads in ICL performance with a BLEU score of 50.13.
For translating Shakespearean to modern English, CoTeX-TB shows marked improvements over both unsupervised and ICL methods, attributed to the superior quality of LLM generations in this specific transfer task.

\begin{figure}[t]
  \begin{center}
\includegraphics[width=\linewidth]{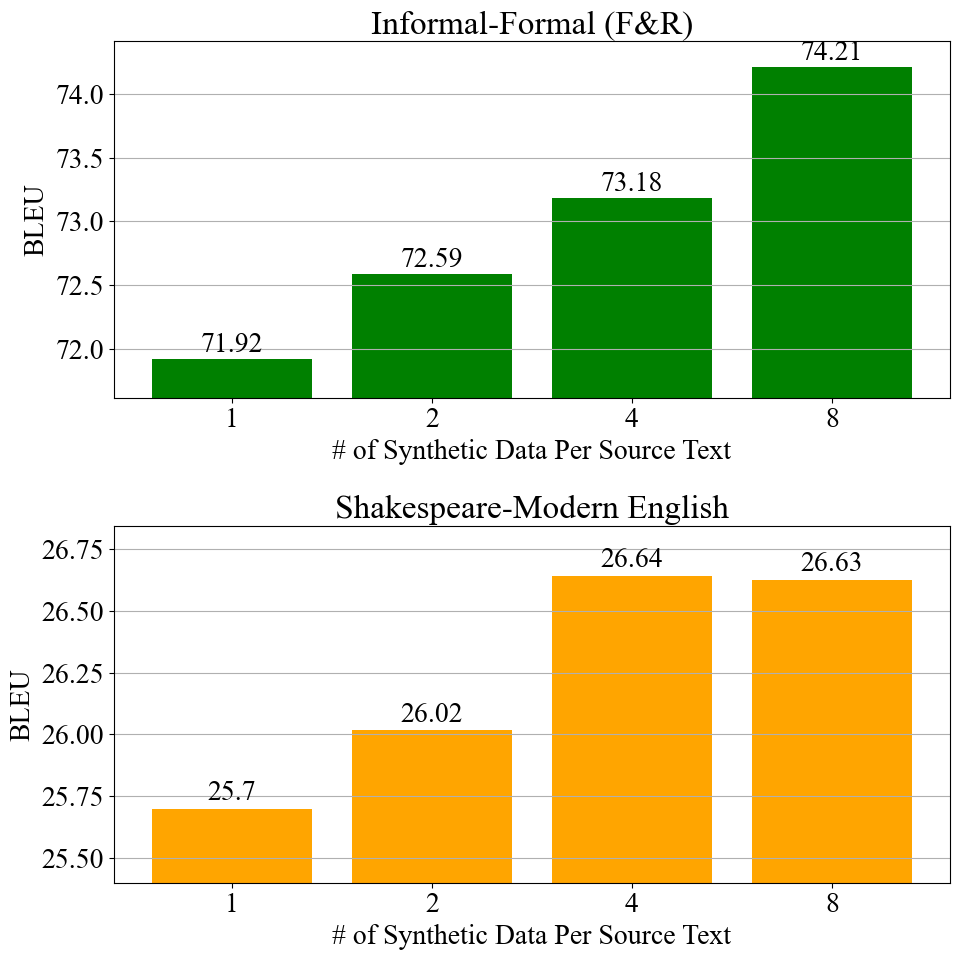}
  \end{center} 
\caption{Scaling up synthetic data per source.}
\label{scale_cot}
\end{figure}

\paragraph{Increasing Synthetic Data per Source Text.} For CoTeX-TB, we conduct an ancillary study to explore the benefits of employing multiple CoT paths with synthetic target texts for a source text. Given a source text $s_i$, the LLM generate $q$ CoT paths, $\{c_{i,1},c_{i,2}, \dots ,c_{i,q}\}$ and their corresponding synthetic target text $\{\hat{t}_{i, 1}, \hat{t}_{i, 2}, \dots , \hat{t}_{i, q}\}$. We select a subset of 5K unique source texts as inputs and investigate the effect of $q$ over a range of $\{2, 4, 8\}$. We experiment with two datasets, Formality (F\&R) and Shakspeare-modern English. Table~\ref{scale_cot} shows a positive correlation between the student model's performance and increasing $q$ values. 


\begin{table}[t]
\centering
\tiny
\begin{tabular}{@{}p{1.3cm}p{5.9cm}@{}}
\toprule
Source Text      & just want to know HAVE YOU BEEN TO THE DOC YET                                                                                                                                                                                                                        \\ \hline
\begin{tabular}[c]{@{}p{1.5cm}@{}}CoTeX\\Generation\end{tabular} & \begin{tabular}[c]{@{}p{5.9cm}@{}}The original text is informal. \\ The use of all caps and the absence of punctuation are informal. The use of ``DOC" is a misspelling of ``doctor". \\ {[}{[}Transferred{]}{]}: I just want to know if you have been to the doctor yet.\end{tabular} \\ \hline
Reference   & I want to know if you have been to the doctor yet.                                                                                                                                                                                                                    \\ \bottomrule\toprule
Source Text      & bembie hit the nail on the head                                                                                                                                                                                                                                       \\ \hline
\begin{tabular}[c]{@{}p{1.5cm}@{}}CoTeX\\Generation\end{tabular} & \begin{tabular}[c]{@{}p{5.9cm}@{}}The original text is informal. \\ The use of ``bembie" is a misspelling of ``bembie". \\ {[}{[}Transferred{]}{]}: Bembie hit the nail on the head.\end{tabular}                                                                                      \\\hline
Reference   & Bembie reached the proper conclusion.                                                                                                                                                                                                                                 \\ \bottomrule
\end{tabular}
\caption{Case study on CoTeX-TB generations.}\label{tab:case_study}
\end{table}

\paragraph{Qualitative Study.}
We now present a qualitative study to delve into rewriting rationales generated by CoTeX-TB. Examples are showcased in Table~\ref{tab:case_study}, derived from Test set of Formality (F\&R) which transfers from informal to formal text. We sort generations by their BLEU scores against gold references and select random high and low-scoring samples. The first example, obtained BLEU of 100, correctly identifies informal components, fixes informal spellings, and yields a formal and grammatical sentence. The second example (BLEU=7.27) misses comprehending the idiom ``hit the nail on the head'' from the source, without translating it into a formal expression. Nevertheless, we note that the LLM (i.e., PaLM2) can appropriately adapt this idiom to ``accurately identified the key point''. This leads us to hypothesize that a smaller LM exhibits potential limitations in its ability to understand implicit style cues. 
\begin{table}[]
\centering
\small
\begin{tabular}{@{}ll@{}}
\toprule

\textbf{Level} & \textbf{Criteria}                                                                                            \\ \midrule
\textbf{Rate A}         & \begin{minipage}[t]{0.8\linewidth}
              \begin{itemize}
              \item  Valid, acceptable and satisfying (subject to the annotator) response;
              \item Accurately identified the most cues for text style transfer;
              \item The reasoning path can directly lead to the transferred text. 
              \end{itemize}
            \end{minipage}  \\ \midrule
\textbf{Rate B}         & \begin{minipage}[t]{0.8\linewidth}
              \begin{itemize}
              \item The response is acceptable but has minor errors that can be improved;
              \item Mirror errors include out-of-context content, minimal factual errors, missing many cues for text style transfer, etc.
              \end{itemize}
            \end{minipage}  \\\midrule
\textbf{Rate C}         &  \begin{minipage}[t]{0.8\linewidth}
                  \begin{itemize}
                  \item The response is relevant but it has significant errors in the content;
                 \item Cannot identify any correct cues for text style transfer.
              \item The reasoning path cannot lead to the transferred text.
              \end{itemize}
              \end{minipage} \\\midrule
\textbf{Rate D}         & \begin{minipage}[t]{0.8\linewidth}
                  \begin{itemize} 
                  \item Invalid and unacceptable response;
                  \item Nothing related to the text style transfer task.
                  \end{itemize}
              \end{minipage} \\\midrule
\multicolumn{2}{l}{\begin{minipage}[t]{0.93\linewidth}\textbf{Instruction:} This task is text styles transfer that transfers a \{\$source\_style\} source text to a target text with style \{\$target\_style\}. Each example includes a source text and the corresponding model-generated rationales of the rewriting process as well as the transferred text. You evaluate the rationales of the rewriting process and do not take the quality of the transferred text into account.\end{minipage}} \\ \bottomrule
\end{tabular}
\caption{Human evaluation protocol and instruction. We adapt the evaluation criteria from~\citet{wu-2023-laminilm} and \citet{wang-2023-selfinstruct}. }\label{tab:human_eval_rubic}
\end{table}
\begin{figure}[t]
    \centering
    \scriptsize
    \begin{tikzpicture}
    \begin{axis}[
      xbar stacked,
      width=0.94\linewidth,
      enlarge x limits={abs=0.2cm},
      enlarge y limits={abs=0.4cm},
      bar width=16pt,
      xmin=0, xmax=50,
      legend style={at={(0.5,-0.2)}, anchor=north,legend columns=-1},
      xlabel={\# of examples},
      symbolic y coords={
        CoTeX-TB-Form.,
        PaLM2-Form.,
        CoTeX-TB-Detox.,
        PaLM2-Detox.,
        },
      ytick=data,
      nodes near coords,
      every node near coord/.append style={font=\scriptsize},
      tick label style={font=\scriptsize},
      y tick label style={rotate=0,anchor=east},
      ]
    \addplot [xbar, fill=green!30, draw=green] coordinates {
    (20,CoTeX-TB-Form.)
    (22,PaLM2-Form.)
    (43,CoTeX-TB-Detox.)
    (45,PaLM2-Detox.)
    };
    \addplot [xbar, fill=blue!30, draw=blue] coordinates {
    (17,CoTeX-TB-Form.)
    (23,PaLM2-Form.)
    (7,CoTeX-TB-Detox.)
    (5,PaLM2-Detox.)
    };
    \addplot [xbar, fill=yellow!30, draw=yellow] coordinates {
    (13,CoTeX-TB-Form.)
    (5,PaLM2-Form.)
    (0,CoTeX-TB-Detox.)
    (0,PaLM2-Detox.)
    };
    \addplot [xbar, fill=red!30, draw=red] coordinates {
    (0,CoTeX-TB-Form.)
    (0,PaLM2-Form.)
    (0,CoTeX-TB-Detox.)
    (0,PaLM2-Detox.)
    };
    \legend{\texttt{Rate-A}, \texttt{Rate-B}, \texttt{Rate-C}, \texttt{Rate-D}}
    \end{axis}
    \end{tikzpicture}
    \caption{
        Human evaluation results of CoT reasoning paths of 50 samples. \textbf{Form.}: formality transfer, \textbf{Detox.}: Detoxification. 
    }\label{fig:human_eval}
    \label{fig:human_eval_user}
\end{figure}
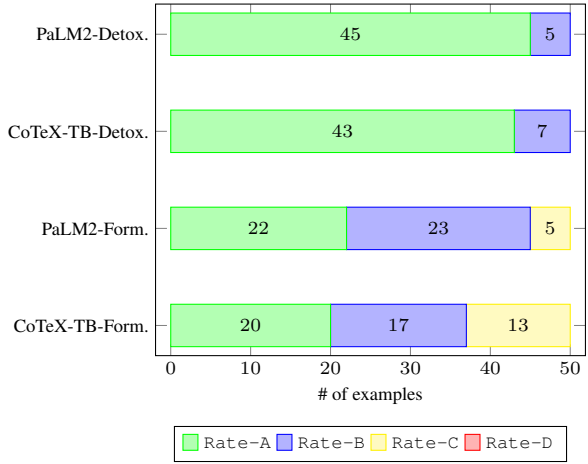



\paragraph{Human Evaluation on Generated Reasonings.}\label{sec:human_eval}
To assess the quality of model-generated rationales (i.e., CoT path) for the rewriting process, we conduct a human evaluation. Following previous works~\cite{wang-2023-selfinstruct, wu-2023-laminilm}, we develop our evaluation protocol and instructions as shown in Table~\ref{tab:human_eval_rubic}. We assemble a team of four human experts to undertake this evaluation. Each annotator was tasked with reviewing 50 generated rationales across different models and transfer tasks. For each evaluation, the dataset provided included the source text, a generated rationale for the rewriting process, and the resultant transferred text. As depicted in Figure~\ref{fig:human_eval}, although CoTeX-TB lags behind the teacher model (PaLM2 Unicorn), 100\% of its responses in the detoxification task and 74\% in the formality transfer task are deemed acceptable. 



\section{Related Work}

When parallel TST datasets are available, numerous studies~\cite{rao-tetreault-2018-dear, shange-2019-semi-supervised, chawla-yang-2020-semi, lai-2021-thank} have utilized a sequence-to-sequence framework for supervised training TST models. To improve model efficacy, multitask learning~\cite{niu-2018-multitask, formality-2019-xu}, lexically constrained decoding~\cite{post-2018-fast}, and task-specific data augmentation~\cite{zhang-2020-parallel, liu-2022-semisupervised} have been incorporated. Addressing the scarcity of parallel data, unsupervised methods have been developed for TST, employing methodologies like disentanglement of latent representations~\cite{liu2020revision, nangi-2021-counterfactuals, yi2021text}, prototype editing~\cite{li-2018-delete}, style rewriting using attribute-specific LMs~\cite{krishna-2020-reformulating}, and reinforcement learning~\cite{luo-2019-dual,hallinan-2023-steer}. Our CoTeX framework explores both parallel and non-parallel data landscapes.
The advent of LLMs has introduced ICL for executing TST with few-shot prompts, bypassing the need for model parameter updates~\cite{reif-2022-recipe, suzgun-etal-2022-prompt}. Yet, these methods typically lack interpretability. In parallel, \citet{saakyan-2023-iclef} employ CoT prompting alongside domain expert feedback to enhance formality transfer and interpretability. Our CoTeX extends to broader range of TST directions, aiming to utilize CoT to provide rewriting explanations and minimize the requirement for human intervention.


\section{Conclusion}
We introduced CoTeX, a novel approach for TST. Through CoT prompting, we elicit the rationals for the style rewriting process from LLMs and then distill both the TST and reasoning capabilities into smaller task-specific models. CoTeX demonstrated its efficiency and effectiveness with and without utilizing parallel data, especially in low-resource scenarios. The CoT reasoning from CoTeX bolstered the explainability of TST models.

\section{Limitations}
 \paragraph{\textbf{TST Directions.}} We incorporate three style transfer directions to enable a clear comparison between target-blind and target-aware CoTeX. Benefiting from the powerful capacity of LLMs, we believe that our method could be extended to a broader array of TST directions (e.g., sentiment transfer). We plan to explore more transfer directions in future work. 

\paragraph{\textbf{Model Selection.}} We only use T5-large as the student model in the paper. We also conduct a concise study to apply CoTeX to the T5-XL model. As results shown in Appendix~\ref{sec:t5xl}, our CoTeX-TA still outperforms SFT on ParaDetox dataset. 

\paragraph{\textbf{Evaluation Metrics.}} Unlike previous studies~\cite{krishna-2020-reformulating, liu-2022-semisupervised}, we abstain from using other automatic metrics (e.g., BERTscore for meaning preservation) to evaluate our models. Our decision is grounded in two main reasons: (1) While these automatic evaluations consider three facets, i.e., preservation of semantic meaning, accuracy of style transfer, and fluency they lack an effective methodology for aggregating these metrics to convey the overall performance~\cite{ostheimer-etal-2023-call}; (2) Our preliminary experiments involving these automatic metrics revealed a misalignment between their outcomes and the BLEU score derived from human-annotated references. We thus opt to report the BLEU score in the paper. Detailed results from our preliminary tests are presented in Appendix~\ref{sec:append_metrics}.

\section{Ethical Consideration}
The primary objective of training CoTeX model is to achieve more computationally efficient and effective models for TST. We focus on the positive TST directions, such as language detoxification. We use an LLM to generate rationales alongside transferred text, which are subsequently distilled into smaller LMs. It's important to acknowledge that the LLM's generation might encompass societal biases~\cite{lucy-bamman-2021-gender} or hallucinations~\cite{zhang-2023-hallucination}, and student models trained with this data could inherit these characteristics of the teacher LLM. Additionally, our CoTeX-TA relies on datasets from prior research. Thus, any biases present in the original annotation processes of these datasets might also be reflected in our trained models. We expect the ongoing work~\cite{ouyang-2022-training,dev-2022-measures} of improving LM's social fairness, faithfulness, and trustworthiness could benefit both teacher and student models.

\bibliography{custom,anthology}

\appendix

\newpage
\textbf{\Large Appendices}






\section{Preliminary Test on Evaluation Metrics}\label{sec:append_metrics}
In our preliminary experiment, we evaluate model performance with several automatic metrics utilized by previous works~\cite{krishna-2020-reformulating, luo-2019-dual, reif-2022-recipe}. These automatic metrics have been widely used in unsupervised TST due to their independence from human-labeled parallel data. However, we find that the outcomes from these metrics do not align with the reference-BLEU score derived from human-annotated references. These automatic metrics evaluate transferred text from three aspects: 
\begin{enumerate}
    \item Similarity: To evaluate the similarity between the source text and the transferred text, we employ BERTscore and self-BLEU. For BERTscore calculations, we use the SimCSE-large model~\cite{gao-2021-simcse} as the backbone. 
    \item Transfer Accuracy: To evaluate the efficacy of the style transfer, we employ a classifier~\cite{babakov-2023-dont} to determine whether the transferred text successfully achieves the desired style.
    \item Fluency: To access the fluency of the transferred text, we compute its perplexity using GPT. Additionally, we utilize a classifier trained on the Corpus of Linguistic Acceptability (CoLA) from \cite{krishna-2020-reformulating} to determine the grammaticality of the transferred text. 
\end{enumerate}

In this preliminary experiment, we conduct experiments using varying training sizes from the formality transfer (F\&R) dataset. These experiments are carried out in a target-blind setting, where we finetune a T5-large model using the synthetic data generated from LLM. For assessing transfer accuracy, we employ a binary classifier introduced by \citet{babakov-2023-dont}, which is a RoBERTa-base model finetuned on the GYAFC's training set. This classifier achieves a test accuracy of 0.91. As Table~\ref{tab:eval_metric} shows, the outcomes from these metrics did not correspond well with the reference-BLEU score. We thus opt to report the BLEU score in the paper. 

\begin{table*}[h]
\centering
\begin{tabular}{@{}rcccccc@{}}
\toprule
\multicolumn{1}{c}{\# of data} & Ref-BLEU & BERTScore & Self-BLEU & Tra. Acc. & PPL   & CoLA \\ \midrule
1000                          & 72.54    & 0.96      & 45.34     & 0.94      & 61.15 & 0.95 \\
2000                          & 73.13    & 0.96      & 46.89     & 0.93      & 59.02 & 0.95 \\
5000                          & 71.92    & 0.96      & 50.71     & 0.90      & 65.54 & 0.94 \\
10000                         & 72.86    & 0.96      & 51.15     & 0.89      & 63.98 & 0.95 \\
20000                         & 72.90    & 0.96      & 49.89     & 0.90      & 64.66 & 0.94 \\ \bottomrule
\end{tabular}
\caption{Preliminary result on GYAFC (F\&R) for investigating evaluation metrics. \textbf{Tra. Acc.}: transfer accuracy, \textbf{PPL}: perplexity. } \label{tab:eval_metric}
\end{table*}

\section{Experiment with T5-XL}\label{sec:t5xl}
We conduct a concise experiment to apply our CoTeX to T5-XL (containing 3B parameters). As Table~\ref{tab:t5xl} shows, our CoTeX-TA outperforms SFT across all the data sizes.
\begin{table}[h!]
\centering
\begin{tabular}{@{}rccc@{}}
\toprule
\multicolumn{1}{l}{\textbf{\# Data}} & \textbf{SFT} & \textbf{CoTex-TB} & \textbf{CoTex-TA} \\ \midrule
1000                                         & 49.15        & 46.64                & \textbf{53.93}     \\
2000                                         & 51.58        & 47.56                & \textbf{54.26}     \\
5000                                         & 52.91        & 47.92                & \textbf{54.83}     \\
10000                                        & 52.32        & 47.96                & \textbf{55.13}     \\
15000                                        & 52.88        & 48.47                & \textbf{55.19}     \\ \bottomrule
\end{tabular}
\caption{Finetuning T5-XL on detoxification dataset with our CoTeX or SFT.}\label{tab:t5xl}
\end{table}



\end{document}